\documentclass[letterpaper, 10 pt, conference]{ieeeconf}  

\IEEEoverridecommandlockouts                              

\makeatletter
\let\NAT@parse\undefined
\makeatother
\usepackage[comma,numbers, compress]{natbib}

\usepackage[hidelinks]{hyperref}
\usepackage{url}
\usepackage{blindtext}
\usepackage{graphicx}
\usepackage[usenames,dvipsnames]{xcolor}
\usepackage[caption=false,font=footnotesize]{subfig}
\usepackage[draft]{fixme}
\fxsetup{theme=color}
\usepackage{soul}
\usepackage{dblfloatfix}
\usepackage{scrextend}
\usepackage{authblk}
\usepackage{amsmath}
\usepackage{amssymb}
\usepackage{booktabs}
\usepackage{soul}
\usepackage{textcomp}
\usepackage{threeparttable}
\usepackage{wrapfig}
\usepackage{multirow}
\usepackage{makecell}

\usepackage{balance}

\usepackage[capitalize]{cleveref}
\crefname{section}{Sec.}{Sec.}
\crefname{table}{Tab.}{Tab.}

\usepackage[tightpage]{preview}

{\noindent\ignorespaces}%
{\par\noindent%
\ignorespacesafterend}
\PreviewEnvironment{maybepreview}

\usepackage{tikz}
\usepackage{pgfplots}
\usepgfplotslibrary{fillbetween}
\usepgfplotslibrary{groupplots}
\usetikzlibrary{arrows}
\usetikzlibrary{positioning,calc}
\usetikzlibrary{decorations.pathreplacing}
\usetikzlibrary{decorations.markings}
\usetikzlibrary{fit}
\usetikzlibrary{shapes.callouts}
\usetikzlibrary{shapes.geometric}
\usetikzlibrary{matrix}
\usetikzlibrary{patterns}

\newcommand{\eg}{e.g.,\ }

\graphicspath{{figures/}}
\usepackage[firstpage=true]{background}
\newcommand\copyrighttext{%
        \parbox{\textwidth}{
                \footnotesize
                Accepted for IEEE Robotics and Automation Magazine (RAM), to appear December 2019.
        }
}

\SetBgContents{\copyrighttext}
\SetBgScale{1}
\SetBgColor{black}
\SetBgAngle{0}
\SetBgOpacity{1}
\SetBgPosition{current page.north}
\SetBgVshift{-0.8cm}

\usepackage[bottom]{footmisc}

\title{\LARGE \bf
Flexible Disaster Response of Tomorrow \\ {\large Final Presentation and Evaluation of the CENTAURO System} 
}

\author[1]{Tobias Klamt\thanks{Author contact: Tobias Klamt \texttt{<klamt@ais.uni-bonn.de>}.\ \ \ \ \ \ \ \ \ 
This work was supported by the European Union's Horizon 2020 Programme under Grant Agreement 644839 (CENTAURO).
}}
\author[1]{Diego Rodriguez}
\author[2]{Lorenzo Baccelliere}
\author[3]{Xi Chen}
\author[4]{Domenico Chiaradia}
\author[5]{Torben Cichon}
\author[4]{\\Massimiliano Gabardi}
\author[2]{Paolo Guria}
\author[6]{Karl Holmquist}
\author[2]{Malgorzata Kamedula}
\author[3]{Hakan Karaoguz}
\author[2]{\\Navvab Kashiri}
\author[2]{Arturo Laurenzi}
\author[1]{Christian Lenz}
\author[4]{Daniele Leonardis}
\author[2]{Enrico Mingo Hoffman}
\author[2,9]{\\Luca Muratore}
\author[1]{Dmytro Pavlichenko}
\author[4]{Francesco Porcini}
\author[2]{Zeyu Ren}
\author[7]{Fabian Schilling}
\author[1]{Max Schwarz}
\author[4]{\\Massimiliano Solazzi}
\author[6]{Michael Felsberg}
\author[4]{Antonio Frisoli}
\author[8]{Michael Gustmann}
\author[3]{Patric Jensfelt}
\author[6]{\\Klas Nordberg}
\author[5]{J{\"u}rgen Ro{\ss}mann}
\author[8]{Uwe S{\"u}ss}
\author[2]{Nikos G. Tsagarakis}
\author[1]{Sven Behnke}

\affil[1]{\footnotesize{Autonomous Intelligent Systems, University of Bonn, Germany}}
\affil[2]{Humanoids and Human-Centred Mechatronics, Istituto Italiano di Tecnologia, Genoa, Italy}
\affil[3]{Department of Robotics, Perception and Learning, KTH Royal Institute of Technology, Stockholm, Sweden}
\affil[4]{PERCRO Laboratory, TeCIP Institute, Sant'Anna School of Advanced Studies, Pisa, Italy}
\affil[5]{Institute for Man-Machine Interaction, RWTH Aachen University, Germany}
\affil[6]{Computer Vision Laboratory, Link{\"o}ping University, Sweden}
\affil[7]{Laboratory of Intelligent Systems, Swiss Federal Institute of Technology (EPFL), Lausanne, Switzerland}
\affil[8]{Kerntechnische Hilfsdienst GmbH, Karlsruhe, Germany}
\affil[9]{School of Electrical and Electronic Engineering, The University of Manchester, United Kingdom}

\begin{document}

\maketitle

\thispagestyle{empty}
\pagestyle{empty}


\noindent

Mobile manipulation robots have high potential to support rescue forces in disaster-response missions.
Despite the difficulties imposed by real-world scenarios,
robots are promising to perform mission tasks from a safe distance.
In the CENTAURO project, we developed a disaster-response system which consists of the highly flexible Centauro robot and suitable control interfaces including an immersive tele-presence suit and support-operator controls on different levels of autonomy.

In this article, we give an overview of the final \mbox{CENTAURO} system.
In particular, we explain several high-level design decisions and how those were derived from requirements and extensive experience of Kerntechnische Hilfsdienst GmbH, Karlsruhe, Germany (KHG)\footnote{KHG is part of the German nuclear disaster-response organization and a potential end-user of the system}.
We focus on components which were recently integrated and report about a systematic evaluation which
 demonstrated system capabilities and revealed valuable insights.

\section{Disaster Response needs Flexible Robots and Intuitive Teleoperation}
\label{sec:approach}

\noindent
Natural and man-made disaster response have increasingly attracted attention in the last decade because of their large potential impact.
One example is the Fukushima disaster in March 2011 in which a Tsunami hit the Japanese coast---including the Fukushima Daiichi nuclear power plant---resulting in a nuclear disaster.
There, rescue workers faced considerable risks to their health \citep{hiraoka2015review}.

Although mobile manipulation robots have much potential to replace humans in such cases, 
the Fukushima nuclear disaster illustrated the gap between robot capabilities and real-world requirements.
In such scenarios, a combination of locomotion and manipulation skills is required.
Locomotion tasks include, \eg energy-efficient traversal of longer distances to approach the disaster site, 
navigation in man-made environments such as corridors, doors, ramps, and staircases, 
as well as overcoming cluttered obstacles or gaps which are caused by the disaster.
Examples for manipulation tasks are the moving of heavy obstacles to clear paths, interactions in made-for-human environments such as (dis)connecting plugs or passing locked doors, as well as using made-for-human electrical tools such as power drills.
Further challenges include: necessary energy resources have to be on board, direct visual contact between an operator and the robot is not available, and maintenance during a mission is not possible---putting the focus on system robustness.

\begin{figure*}\includegraphics{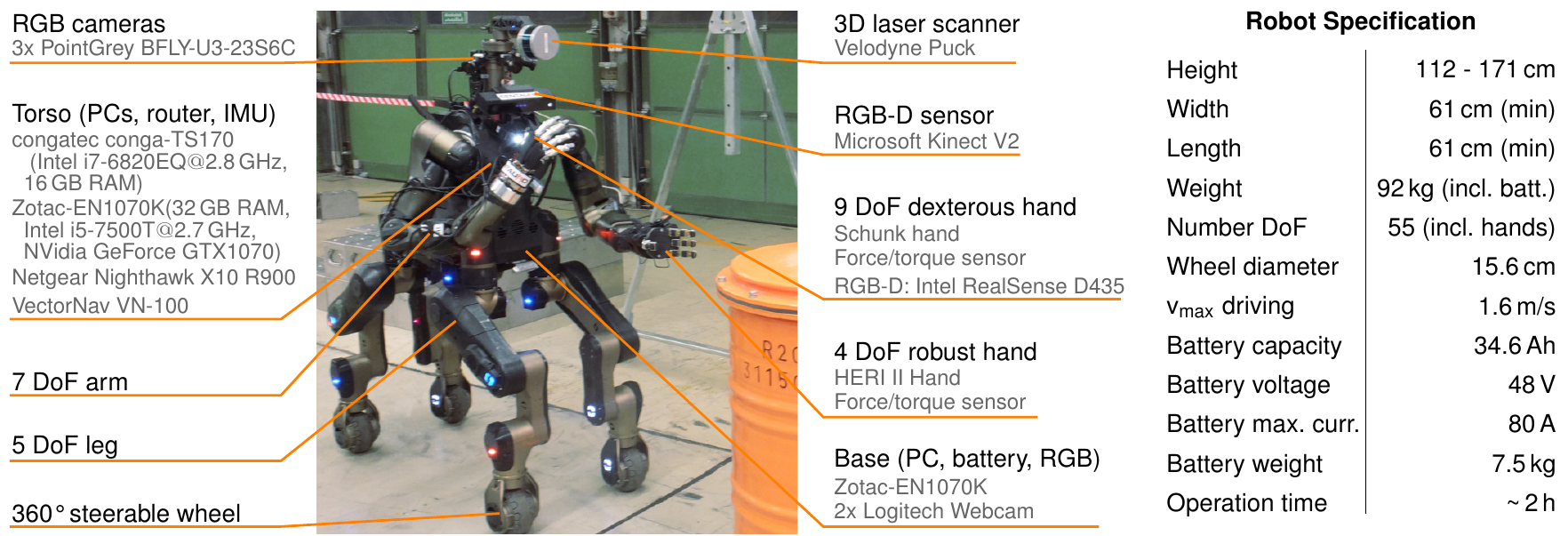}
	\vspace{-0.6cm}
	\caption{The Centauro robot. \textit{Left:} Centauro approaching a cask to conduct a radiation measurement. 
		Labels describe the hardware components and their location. \textit{Right:} robot hardware specifications.}
	\label{fig:centauro_robot}
	\vspace{-0.5cm}
\end{figure*}

In recent years, continuous developments in actuation, sensing, computational hardware, and batteries enabled the development of robotic platforms which provide essential features like flexibility, precision, strength, robustness, energy efficiency, and compactness such that robots can be deployed in environments designed for humans.
In the 2015 DARPA Robotics Challenge, 
teams were required to solve a set of mostly known locomotion and manipulation tasks and demonstrated the state-of-the-art for applied mobile manipulation robots.
Successful robot entries include DRC-HUBO, Momaro, CHIMP, and RoboSimian~\citep{zucker2015general, schwarz2017nimbro, stentz2015chimp, hebert2015mobile}.
More recent mobile manipulation robots are, e.g., RoboMantis~\citep{motiv2019robomantis} and Spot~\citep{boston2019spot} which possess flexible kinematic capabilities but no methods to control these platforms have been presented.
The humanoid Toyota T-HR3 \cite{toyota2017toyota} and the WALK-MAN robot attached to a wheeled base~\citep{negrello2018humanoids} have been controlled through a head-mounted display (HMD) and by mirroring user movements to the robot.
In the case of T-HR3, additional force feedback is available.
However, evaluation of these systems only addressed tasks of limited difficulty which were known in advance such that operator training was possible.

The CENTAURO project aimed at increasing the applicability of mobile manipulation robots to real-world missions.
Hence, the development focused on a system that is flexible enough to solve a wide range of considerably more challenging and realistic tasks, compared to what has been presented in related works.
This includes task accomplishment without previous specific training of the operators---as in realistic disaster-response missions.
We are convinced that a single operator interface that is flexible to the considered wide range of tasks---while being easy to use---does not exist.
Instead, a set of operator interfaces with complementary strengths and different degrees of autonomy is proposed.
For tasks of limited complexity, high-level operator inputs suffice to control autonomous functionalities while relieving cognitive load from the operator.
For rather difficult tasks, human experience and capabilities for scene understanding and situation assessment are indispensable.
Thus, it is highly desirable to transfer these cognitive capabilities into the disaster scene, through, e.g., the developed tele-presence suit.

Many technical details of the systems have been already described in~\citep{klamt2019remote}---a report on the intermediate system status one year before the end of the project.
In this article, we present the final system at the end of the project.
A focus is put on describing how several fundamental design decisions were motivated.
Furthermore, we describe components that have been recently developed and, thus, are not included in the before-mentioned report, such as a new robust end-effector, a new motion controller, a bi-manual control interface for the tele-presence suit, bi-manual autonomous grasping, a terrain classification method for autonomous locomotion, and virtual models for the robot and its environments.
Finally, a systematic evaluation is presented which goes significantly beyond previously reported experiments.
It consists of considerably more challenging tasks to allow for an assessment of the real-world applicability of the system.

\section{The Centauro Robot}
\label{sec:robot}

\noindent
The overall robot design is motivated by the requirements of typical disaster-response missions and experiences of KHG,
who operates a fleet of disaster-response robots.
Those platforms are mostly tracked vehicles with a single arm ending in a 1-DoF (Degrees of Freedom) gripper.
Moreover, additional monochromatic or RGB cameras provide scene visualizations to the operator.

\subsection{Hardware Description}
Regarding the locomotion concept, KHG reported that tracked approaches are robust and easy to control but exhibit difficulties in very challenging environments such as large debris or gaps.
Moreover, the restricted energy efficiency of tracks requires the systems to have large batteries---increasing weight.
Better energy efficiency is generally provided by wheeled platforms but their mobility is limited to sufficiently flat terrains.
On the other hand, legged robots are promising to traverse challenging terrains (including steps and gaps) since only isolated footholds are needed.
For the Centauro robot (see \cref{fig:centauro_robot}), we chose a hybrid driving-stepping design with legs ending in wheels.
In this manner, the complementary strengths of the two locomotion approaches are combined.
Furthermore, the robot can switch between locomotion modes without posture changes which enables additional unique motions such as individual feet movement relative to the base while keeping ground contact---a valuable property for balancing.
Wheels are 360\textdegree\ steerable and actively driven allowing for omnidirectional driving.
Legs incorporate five DoFs each, with three upper joints arranged in a yaw-pitch-pitch configuration and two lower joints for wheel orientation and steering.
When stepping, wheel rotation is blocked such that the grip is similar to point-like contacts of legged robots.
We chose a design with four legs due to its higher static stability (support polygon area), compared to bipedal robots.

KHG also reported that bi-manual setups can be advantageous in several situations but are rarely present on the market.
Furthermore, operators sometimes experience problems in understanding kinematic constraints of the manipulator arms.
Motivated by these observations, we chose an anthropomorphic upper-body design for Centauro resulting in an eponymous centaur-like body plan.
It enables effective, bi-manual manipulation in human workspaces.
A yaw joint connects the upper body to the lower body.
The two 7-DoF robot arms possess a quasi-anthropomorphic profile providing an intuitive kinematic system understanding for the operators.
To obtain versatile grasping capabilities, two different end-effectors with complementary strengths are incorporated.
On the right, a commercial, anthropomorphic, 9-DoF Schunk SVH hand provides dexterous grasping capabilities.
On the left, a flexible, 4-DoF HERI II hand \citep{RenHERIIIRobust2018} allows for higher payload and exhibits higher compliance and robustness.
The HERI II hand was particularly developed to overcome the limitation of typical under-actuated hands in executing basic dexterous motions such as pinching and triggering tools, while maintaining the main advantages of under-actuated designs.

With respect to the sensor setup, 
KHG reported that the operation of their robots, based solely on camera images, 
puts a high cognitive load on the operators and requires extensive training to derive correct situation assessments from those images.
The lack of assessment of the overall kinematic system state and of the robot's environment cause uncertainty and stress.
Therefore, Centauro incorporates a multi-model sensor setup which provides comprehensive visualizations to the operators and input to (semi)autonomous control functionalities.
It includes a 3D laser-scanner with spherical field-of-view, an RGB-D sensor, and three RGB cameras in a panoramic configuration at the robot head.
A pan-tilt mechanism allows for adaptation of the \mbox{RGB-D} sensor pose---facilitating manipulation at different locations.
Furthermore, an inertial measurement unit (IMU) is mounted in the torso.
Two RGB cameras under the robot base provide views on the feet.
An RGB-D sensor at the right wrist creates an additional perspective during manipulation.
Finally, two 6-DoF force/torque sensors are mounted between the robot arms and hands.

To the best of our knowledge, there is no system on the market that provides hybrid locomotion combined with bi-manual manipulation and the required sensor setup. 
The robot has been mainly fabricated of an aluminum alloy.
To drive the robot, five sizes of torque-controlled actuators have been developed.
The compactly integrated series elasticities provide mechanical robustness to impacts, and torque sensing via deformation monitoring.
Based on an effort analysis, a suitable actuator was chosen for each joint~\cite{kashiri2019centauro}.
In addition, the robot body houses a Wi-Fi router, a battery, and three PCs for real-time control, perception, and higher-level behaviors.

One property of the considered disaster-response domain is radiation which requires robot hardening.
This can be realized through either massive lead cases for the computational hardware---adding significant weight---
or the use of radiation-immune compute hardware that is severely limited, compared to commercial off-the-shelf (COTS) hardware.
We developed the CENTAURO system as a scientific demonstrator that aims at pushing the state-of-the-art.
Hardening the system against radiation would have affected our research by causing considerable limitations to the modular hardware design and the cost-efficient use of COTS components.
We thus see system hardening as a task for further development towards an end-user product. 

\subsection{Software Description}
For low-level hardware communication, we developed \textit{XBotCore}, an open-source,
real-time, cross-robot software platform~\citep{MURATORE_XBOT_JOURNAL:_2017}.
It allows to seamlessly program and control any robotic platform, providing a canonical abstraction that hides the specifics of the underlying hardware.
Higher-level motion commands are processed in the developed \emph{CartesI/O} motion controller~\citep{laurenzi2019cartesio} which can consider multiple \emph{tasks} and \emph{constraints} with given, situation-specific priorities.
The controller resolves a cascade of QP problems, 
each one associated to a priority level, 
taking into account the optimality reached in all the previous priority levels.
The high computational cost of the successive QP problems is alleviated by assessing our IK loop in a range between 1000 and 250 Hz. 


\section{The Operator Station}
\begin{figure}[b!]
	\centering
	\vspace{-0.4cm}\includegraphics{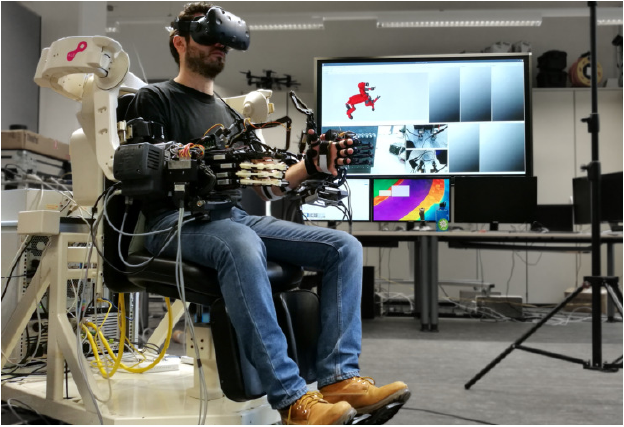}
	\caption{The tele-presence suit allows the operator to intuitively control the remote Centauro robot in bi-manual manipulation tasks while providing force feedback. The support operator station can be seen in the background.}
	\label{fig:telepresence_suit}
\end{figure}

\begin{figure*}
	\vspace{-0.2cm}
	\centering\includegraphics{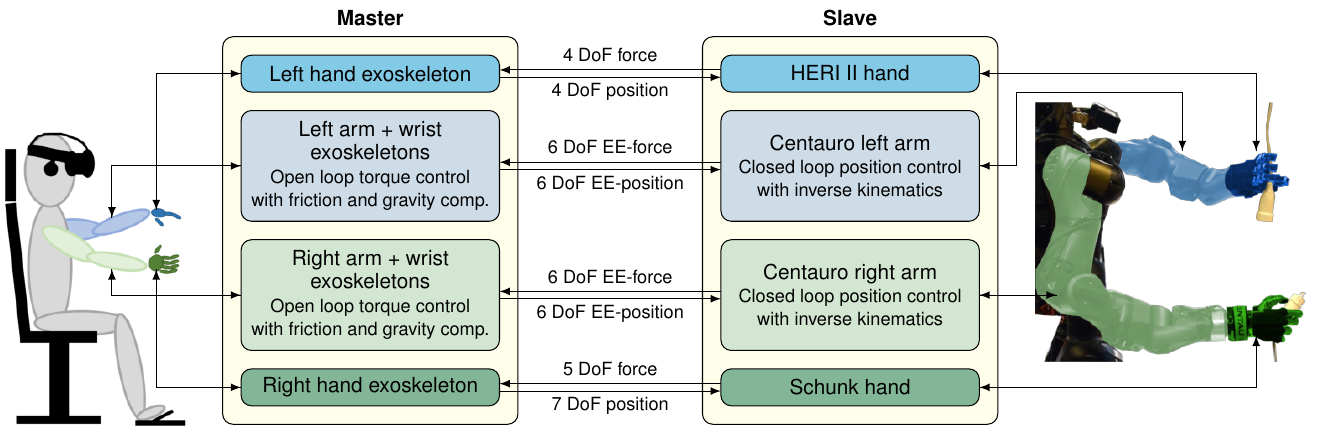}
	\caption{Multilateral position-force teleoperation architecture. While hand positions are perceived on joint level and hand forces are measured for each finger, arm positions and forces are projected to the end-effectors.}
	\vspace{-0.4cm}
	\label{fig:bimanualControlSchema}
\end{figure*}
\noindent
KHG reported that during their real-world missions, 
human concentration suffers from a high cognitive load and thus operators are usually exchanged approximately every 15 minutes.
The considerably more complex kinematics of Centauro would probably even increase this effect.
In discussion with KHG, we identified several reasons for these high cognitive loads:
1)~camera images are not sufficient to generate comprehensive situation understanding---especially if environment contact has to be assessed, 2)~operators have problems to understand the robot kinematics if they cannot see them and cannot assess joint limits and self-collisions, 3)~the continuous direct control of all robot capabilities requires an exhausting, continuous high concentration level.

We derived several requirements for the CENTAURO operator station from these experiences.
Regarding visualization, we aimed at providing a more intuitive situation understanding through a 3D virtual model of the robot and its environment which is generated from joint states, the IMU, the laser scanner, and RGB-D sensors.
It is enriched with camera images from several perspectives.
We refer to these digital counterparts of the robot and its environment as \textit{digital twins}.
Moreover, we added force feedback to increase the situation assessment, especially if environment contacts are involved.
Regarding control, 
several interfaces with different strengths were implemented.
The main operator intuitively controls the robot through a full-body tele-presence suit,
while support operators are able to command additional control functionalities on different levels of autonomy.

\subsection{Full-body Tele-presence Suit}

\begin{figure*}
	\vspace{-0.4cm}
	\centering\includegraphics{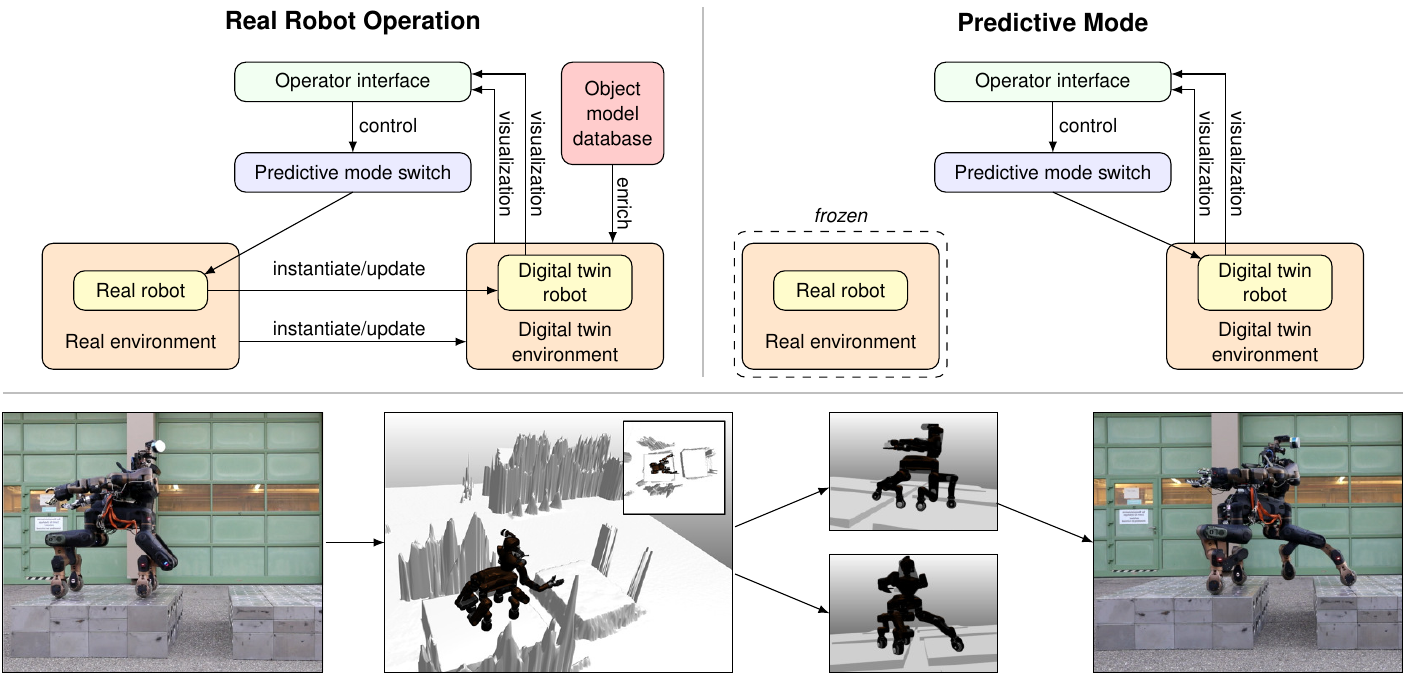}
	\caption{Utilizing digital twins in predictive mode. \emph{Top:} communication architecture for real robot operation and the predictive mode. \emph{Bottom:} when Centauro arrives at the gap, the digital twins of the robot and its environment represent the same situation. The operators switch to the predictive mode and try out different strategies to overcome the gap while the real robot is frozen. Finally, the identified solution can be executed on the real robot.}
	\vspace{-0.5cm}
	\label{fig:digital_twin_overview}
\end{figure*}

The main part of the full-body tele-presence suit is the upper-body exoskeleton with arm exoskeletons covering shoulders and elbows, two spherical wrist exoskeletons, and two hand exoskeletons (\cref{fig:telepresence_suit}).
Hence, complex operator arm, wrist, and hand motions can be intuitively transferred to the robot.
While the arm motions can be also realized through off-the-shelf lightweight manipulators,
the exoskeleton has been designed to maximize performance in human-robot interaction, including high transparency, large workspace and high maximum payload/rendered force.
In contrast to similar teleoperation systems~\cite{Mallwitz2015THECA},
the presented exoskeleton considers force-feedback of both the arm and the hand-palm. 
The arm exoskeletons cover about 90\% of the natural workspace of the human upper arms without singularities.
To keep the number of moving masses low, the device is driven by electric actuators remotely located behind the operator seat and connected to the joints through idle pulleys and in-tension metallic tendons.
Force feedback is provided for each individual joint.
Two rotational \mbox{3-DoF} forearm-wrist exoskeletons track operator wrist motions and provide independent torque feedback for each axis~\cite{buongiorno2018wres}.
Their development focused on the convenience to be worn and an open structure in order to avoid collision between parts during bi-manual operational tasks.
The 6D end-effector poses of the two arms are computed and sent to the Centauro robot which uses them as references to compute joint values through inverse kinematics.
Two hand exoskeleton modules track operator finger motions and provide grasping force feedback~\cite{gabardi2018design}. 
The overall architecture of the tele-presence suit is depicted in \cref{fig:bimanualControlSchema}.
The control incorporates friction and gravity compensation as well as the \textit{time domain passivity approach} combined with a position-force scheme to achieve high transparency~\cite{BuongiornoEtAl2019}.

Since Centauro's quadrupedal lower-body is considerably different to the bipedal human lower-body, direct control of all four legs through an exoskeleton is not feasible.
Instead, we decided to use pedals to control omnidirectional driving while more challenging locomotion is controlled by the support operators.
In addition, the operator wears an HMD for immersive 3D visualizations.

\subsection{Support Operator Interfaces}

\noindent
While the tele-presence suit is advantageous for challenging manipulation, several other tasks can be solved with an increasing level of autonomy---requiring less input from the operator.
Omnidirectional driving can be controlled through a joystick. 
A 6D mouse provides Cartesian control of 6D end-effector poses for arms and legs. 
For more complex motions, a keyframe editor enables configuration control and motion generation in joint space and Cartesian end-effector space. 
A semi-autonomous stepping controller is used for walking across irregular terrain.
The operator can trigger individual stepping motions while the controller automatically generates a statically stable robot configuration and a subsequent foot stepping motion sequence, including ground contact estimation.
The operator can also define goal poses for an autonomous driving-stepping locomotion planner \mbox{(\cref{sec:auto_loco})}.
Bi-manual manipulation can be controlled through the 6D mouse interface, through the keyframe editor, or by defining targets for autonomous grasping (\cref{sec:auto_manip}).

\subsection{Situation Prediction with Digital Twins}

\noindent
A digital twin virtually represents a real object including its physical and kinematic properties, appearance, data sources, internal functions, and communication interfaces.
We utilize the \emph{VEROSIM} simulation environment
for the digital twin (in-the-loop with the real robot).

In addition, the robot can be placed in simulated environments to support development.
Virtual sensor models can generate detailed measurements of the virtual environment and physical interactions of rigid bodies are simulated.
The digital twin interfaces to external control and processing units are identical to the real robot~\cite{cichon2016sbs}.

The robot's digital twin is accompanied by a digital twin of the robot environment.
3D meshes are generated from laser scanner measurements and can be overlaid by texture coming from, \eg RGB cameras.
A large model database contains numerous static objects like steps or plugs and dynamic objects like valves or doors. 
If such objects are recognized, virtual models of these objects can automatically be positioned in the virtual environment.
Hence, limitations of sensory input, which are caused, \eg by occlusion, are mitigated resulting in a holistic, detailed virtual representation of the scene.
This representation is used to generate immersive user visualizations both for 3D HMDs and 2D renderings from arbitrary view poses~\cite{cichon2016operatorinterface}.
Visualizations can be enriched with control interface specific data such as locomotion paths, which can be inspected before execution.

During robot operation, challenging situations might occur for which the operators are uncertain about the optimal solution.
Since the digital twins of the robot and its environment accompany the mission and are frequently updated, the same situation is represented in the virtual environment.
A predictive mode is activated by freezing the real robot in its current configuration and decoupling the digital twins from their real counterparts. 
Subsequently, the operator control is redirected from the real robot to its digital twin, as visualized in \cref{fig:digital_twin_overview}.
Consequently, operators interact with the virtual representation as they would do with the real robot.
This can be used to, \eg evaluate multiple approaches to solve a specific task.
Once a satisfying solution is found, the task can be executed with the real robot~\cite{cichon2018icarcv}.

\begin{figure*}
	\centering
	\vspace{-0.4cm}\includegraphics{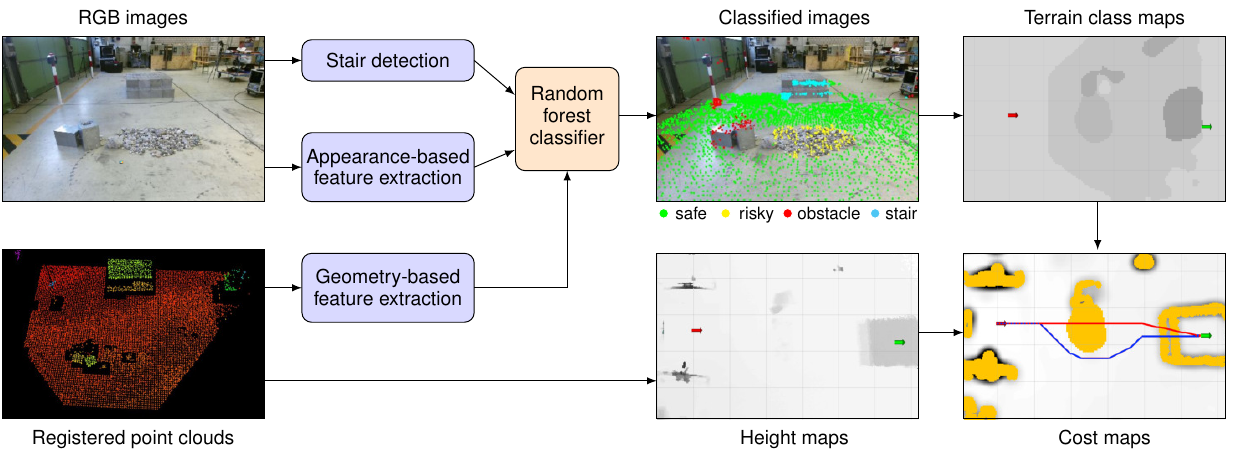}
	\vspace{-0.4cm}
	\caption{Autonomous locomotion cost map generation. Point clouds are processed to height maps to finally generate cost maps. This pipeline is enriched by parallel terrain class computation. Geometry-based and appearance-based features are extracted from point clouds and RGB images. Stairs are detected in an additional module. Outputs are merged in a random forest classifier. 
		Terrain classes are added to the cost computation. 
		Note that the patch of gravel in front of the robot (red arrow) is not recognized well in the height map. The terrain class map clearly represents this area and enriches the cost map with this information. The cost map also shows two generated paths to a goal pose on top of the staircase (green arrow) from which the operator can choose. The red path represents a low weight for terrain class-based costs---resulting in driving over the patch of gravel. The blue path incorporates the terrain class-based cost term with a higher weight---the robot avoids the gravel.}
	\vspace{-0.3cm}
	\label{fig:auto_loco_pipeline}
\end{figure*}

\section{Operator Relief through Autonomous Control Functions}
\label{sec:autonomy}

\noindent
While intuitive teleoperation is superior for solving complex, unknown tasks, disaster-response missions also include several tasks which can be automated in order to reduce the high cognitive load on the operators, e.g., navigation to a desired location or grasping a tool.
Moreover, in comparison to direct teleoperation, autonomy is often faster and depends less on a reliable data link. 
However, the development of autonomous skills is challenging due to the level of variability of the environment and the tasks.

\subsection{Autonomous Hybrid Driving-Stepping Locomotion}
\label{sec:auto_loco}

\noindent
The many DoF that have to be controlled combined with the required precision and balance assessment would put a high cognitive load on the operator and might result in ineffective slow locomotion.
We developed a locomotion planner that combines omnidirectional driving and stepping capabilities in a single graph search-based planning problem~\cite{klamt2017anytime}.
The environment is represented with 2D height maps computed from laser scanner measurements.
These maps are processed to costs for the robot base and individual feet enabling precise planning.

2D height maps are insufficient to represent terrains such as fields of small gravel.
Therefore, the planning pipeline is enriched with terrain classification providing additional semantics (\cref{fig:auto_loco_pipeline}). 
A geometry-based analysis derives the terrain slope and roughness from point clouds.
In parallel, a vision-based terrain classification is performed on RGB images by employing a convolutional encoder-decoder neural network.
Initial training on the CityScapes dataset~\cite{cityscapes} provides the network reliable representations of drivable surfaces, walls, vegetation, and terrain.
Subsequent training on custom datasets from forests and buildings refines the terrain representations.
Focus was put on the classification of staircases which have many similarities to obstacles but differ in traversability.
Finally, all features are fused to pixel-wise traversability assessments (safe/risky/obstacle/stair) which are incorporated in the planner as an additional cost term.

The fine planning resolution (2.5\,cm) and high-dimensional (7\,DoF) robot representation result in rapidly growing state spaces that exceed feasible computation times and available memory for larger planning problems.
We improved the planner to generate an additional coarse, low-dimensional (3\,DoF), and semantically enriched, abstract planning representation~\citep{klamt2019learning}.
The cost function of this representation is learned by a convolutional neural network that is trained on artificial, short planning tasks but generalizes well to real-world scenes.
An informed heuristic (containing knowledge about the environment and the robot) employs this abstract representation and exploits it to effectively guide the planner towards the goal.
This accelerates planning by multiple orders of magnitude with comparable result quality.

\subsection{Autonomous Dual-Arm Manipulation}
\label{sec:auto_manip}

\noindent
While object grasping could be executed through the tele-presence suit, 
autonomous grasping is promising to relieve cognitive load from the operator.
Several objects, e.g., tools, require functional grasps---they must be grasped in a specific way to enable their use, which pose challenges to the autonomy.
Additional challenges arise for bi-manual tasks, which force the execution of collision-free synchronous grasping motions.
\begin{figure}
	\centering
	\scalebox{0.5}{\includegraphics{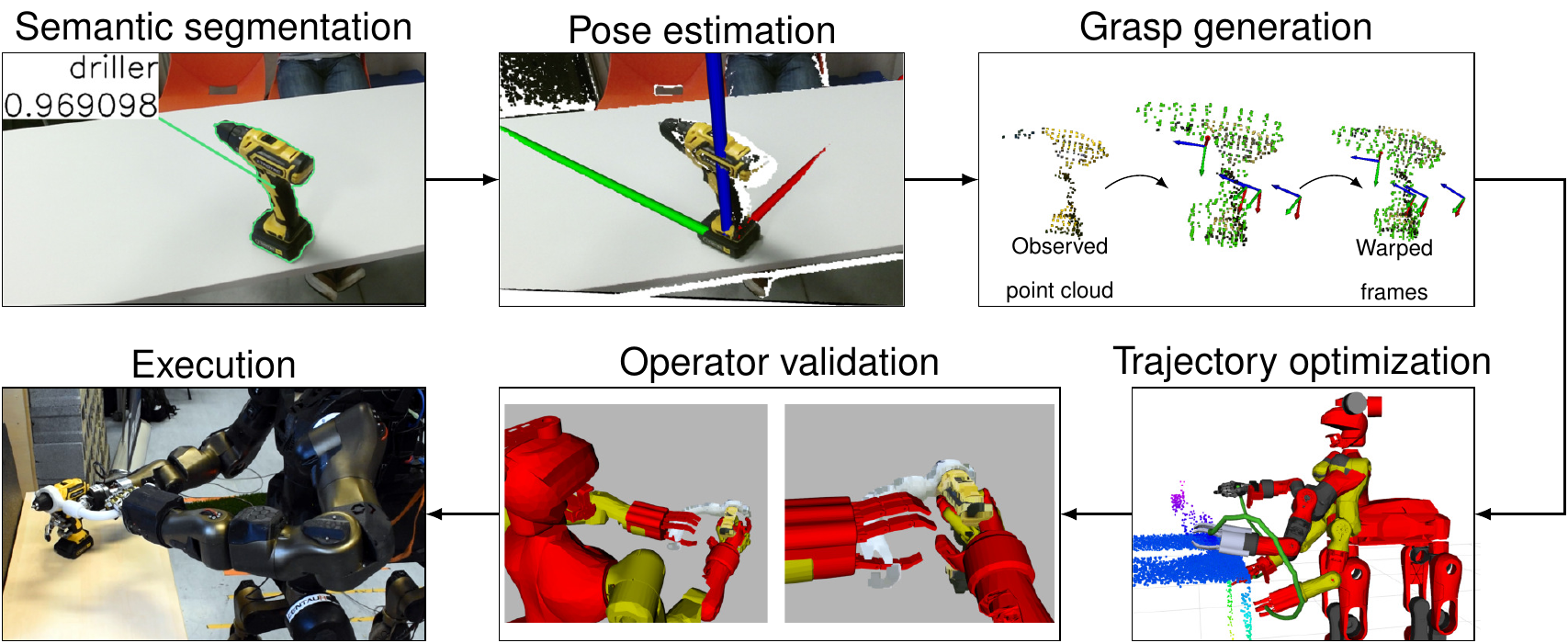}}
	\caption{Autonomous manipulation pipeline. From RGB-D images perceived by the robot, 
		the semantic segmentation finds the contour of the object, e.g. a drill,
		and extracts the object point cloud.
		Based on the contour and the point cloud, the 6D pose of the object is estimated.
		Subsequently, the segmented point cloud is registered non-rigidly using the shape space of the object category and control poses (e.g., pre-grasp and grasp poses) for both arms are warped from the grasping motion of the canonical object.
		The final joint trajectory is generated by a trajectory optimizer that guarantees collision-free motions.
		The resulting motion is verified by the operator and finally executed.
	}
	\vspace{-0.6cm}
	\label{fig:manipulation_pipeline}
\end{figure}

The presented autonomous grasping pipeline enables bi-manual functional grasping of unknown objects if other instances of this object class are known.
Input is RGB-D data.
First, semantic segmentation---based on Refine-Net with additional post-processing---finds object contours and outputs corresponding point clouds.
It is trained on synthetic scenes with objects coming from turn table captures or CAD mesh renderings.
Subsequently, the object pose is estimated from the extracted point cloud and a grasp is generated.
Grasp generation is based on a non-rigid registration that incorporates class-specific information embedded in a low-dimensional shape space for the object category.
To learn the shape space of a category, object point clouds are registered towards a single canonical instance of the category~\cite{Rodriguez2018a,Rodriguez2018b}.
During inference, we optimize a shape that matches best the observed point cloud. 
Finally, individual grasping motions for each arm, which are known for the canonical model, are warped to fit the inferred object shape.
Arm trajectories are subsequently optimized by incorporating collision avoidance, joint limits, end-effector orientation, joint torques, and trajectory duration~\cite{Pavlichenko2018}.
Since the optimization is performed in joint space, multiple end-effectors can be easily added to the optimization problem by only incorporating additional terms in the cost function.
For manipulating an object (once it is grasped) with multiple end-effectors simultaneously, an additional term that penalizes violations of the closed kinematics chain constraint is added.
The planned motion is finally presented to the operator for verification.
The execution of the complete planning pipeline takes less than seven seconds.

\section{Evaluation}
\label{sec:evaluation}

\begin{table}[t!]
	\centering
	\caption{Results of the KHG evaluation.}
	\label{tab:results}
	\begin{tabular}{l|c|c}
		\toprule
		\makecell[c]{\textbf{Task}} & \makecell[c]{\textbf{Rating}\\\footnotesize{\textbf{[\%]}}} & \makecell[c]{\textbf{Time}\\\footnotesize{\textbf{[mm:ss]}}}\\ 
		\toprule
		\multicolumn{3}{c}{\textbf{Locomotion Tasks}}  \\ \toprule
		\makecell[l]{\textit{\textbf{Step Field}}: Navigate over an uneven step\\ field with random debris on it.} & 100 & 87:38\\ \midrule
		\makecell[l]{\textit{\textbf{Staircase}}: Climb a flight of stairs.} & 100 & 10:02\\ \midrule
		\makecell[l]{\textit{\textbf{Gap}}: Overcome a 50\,cm gap.} & 100 & 03:26 \\ 
		\toprule
		\multicolumn{3}{c}{\textbf{Manipulation Tasks}}  \\ \toprule
		\makecell[l]{\textit{\textbf{Fire Hose}}: (Dis-)connect a fire hose to a\\nozzle using a special wrench.} & 100 & 10:10\\ \midrule
		\makecell[l]{\textit{\textbf{230\,V Connector (Standard)}}: Plug a standard\\ 230\,V plug into a cable outlet.} & 100 & 06:50\\ \midrule
		\makecell[l]{\textit{\textbf{230\,V Connector (CEE)}}: Plug a CEE-type\\ plug into an outlet with lid.} & 100 & 10:00\\ \midrule
		\makecell[l]{\textit{\textbf{Shackle}}: Fix a shackle to a metal ring at the\\ wall by inserting and turning a screw.}	& 100 & 24:35\\ \midrule
		\makecell[l]{\textit{\textbf{Electrical Screw Driver}}: Screw a wooden\\ board (held with one hand) to a wooden block\\ with an electrical screw driver.} & 100 & 06:21 \\ \midrule
		\makecell[l]{\textit{\textbf{Power Drill}}: Use a two-handed power drill to \\make holes at marked places in a wooden wall.} & 90 & 02:50\\ 
		\toprule
		\multicolumn{3}{c}{\textbf{Combined Tasks}}  \\ \toprule
		\makecell[l]{\textit{\textbf{Door}}: Unlock and open a regular sized\\door with a handle and pass it.} & 100 & 13:30 \\ \midrule
		\makecell[l]{\textit{\textbf{Platform\,+\,Gate-/ Lever-type Valve}}: Approach\\ and climb a platform with the front feet and\\ open and close a gate-/lever- type valve.} & \makecell[c]{60/\\100} & \makecell[c]{23:30/\\06:50} \\ \midrule
		\makecell[l]{\textit{\textbf{Grasp / Visualize Pipe Star }}: Grasp/visualize\\ defined positions of a pipe star on the ground.} & \makecell[c]{95/\\100} & \makecell{14:25/\\06:09} \\ 
		\toprule
		\multicolumn{3}{c}{\textbf{Autonomous Tasks}}  \\ \toprule
		\makecell[l]{\textit{\textbf{Autonomous Locomotion}}: Autonomously\\navigate around debris and climb stairs\\to reach a goal pose on an elevated platform.} & 75 & 02:50\\ \midrule
		\makecell[l]{\textit{\textbf{Autonomous Manipulation}}: Autonomously\\grasp an unknown, two-handed power drill.} & 75 & 00:46 \\ \bottomrule		
	\end{tabular}
	\vspace{-0.5cm}
\end{table}

\noindent
The final CENTAURO system was evaluated in a systematic benchmark at the facilities of KHG. 
Tasks were designed based on KHG's knowledge about real-world disaster-response missions.
While the intermediate evaluation reported in~\cite{klamt2019remote} was carried out at the same facilities, one year of further, intensive system development significantly extended and advanced the CENTAURO system.
Hence, the described tasks in this article are designed to be considerably more challenging to demonstrate the significantly increased real-world applicability of the final CENTAURO system.

All tasks were performed without previous training and with the operator station located in a separated room---preventing direct visual contact.
The robot operated mostly untethered relying on power supply through its battery and a wireless data link.
A neutral referee rated the goal achievement of the task and ensured that every form of communication between persons at the robot location and the operators was prohibited.
Since the system had research demonstrator status and this evaluation was the first time all required components were integrated into a disaster-response system, we permitted up to three runs for each task of which only the best run was rated.
Nevertheless, most tasks were conducted successfully at the first trial without further runs.

The tasks are not described in chronological order but in categories.
During the evaluation, the \emph{Platform\,+\,Lever} task was one of the first manipulation tasks to be performed. 
Unfortunately, a mechanical part in the upper-body exoskeleton broke during this task and we were not able to repair it in the limited time of the evaluation meeting.
Therefore, we opted to solve all remaining tasks with the remaining interfaces.
However, at the end of this chapter, we present an isolated evaluation of the tele-presence suit.

The task specific performance of the evaluation at KHG is summarized in~\cref{tab:results}. 
Impressions of the experiments are given in~\cref{fig:experiments}.
Further video footage is available online\footnote{\href{https://www.ais.uni-bonn.de/videos/RAM_2019_Centauro}{\scriptsize{\url{https://www.ais.uni-bonn.de/videos/RAM_2019_Centauro}}}}.

\subsection{Locomotion Tasks}

\noindent
Locomotion capabilities were evaluated in three tasks.
In the \emph{Step Field} task, the robot had to traverse a step field which was built from concrete blocks in a random grid scheme. 
The task difficulty was considerably increased through wooden bars which were randomly placed on the step field.
We used the semi-autonomous stepping controller and the joystick for omnidirectional driving to successfully solve this task.
Manual adaptations through the 6D mouse and the keyframe editor were frequently required, due to the irregular ground structure caused by the wooden bars.
In addition, the RGB camera under the robot base providing a detailed view on the rear feet essential for situation assessment, lost connection. 
This issue was compensated by using the right arm RGB-D sensor when executing individual rear foot motions.

In the \emph{Staircase} task, the robot had to climb a staircase with three steps.
We used predefined motion primitives and joystick driving control to solve this task.
Once the robot was completely on the staircase, motions could be triggered in a repetitive manner.
Hence, climbing longer staircases would only require more repetitions of these motions.
In the \emph{Gap} task, the robot had to overcome a 50\,cm gap.
We used motion primitives and joystick commands to successfully solve this task.
In addition, the predictive mode of the digital twin was employed to evaluate different strategies (see \cref{fig:digital_twin_overview}).
The latter has been demonstrated isolated and is not included in the measured time.

\subsection{Manipulation Tasks}

\noindent
Disaster-response manipulation capabilities were evaluated in six tasks.
In the \emph{Fire Hose} task, the robot had to connect and to disconnect a fire hose to a nozzle.
An integrated sealing caused high friction such that a designated wrench was used.
Dexterous grasping capabilities of the Schunk hand were required to mount the fire hose to the nozzle.
The \mbox{HERI II} hand operated the wrench. 
This task was accomplished using the 6D mouse control.
Its feature to enable only certain motion axes facilitated precise alignment.

In the \emph{230\,V Connector (Standard)} task, the robot had to connect and to disconnect a standard 230\,V plug placed in one hand to a cable power outlet hanging from the ceiling.
After grasping the power outlet, precise alignment was challenging since vision was hindered through the enclosure of the power outlet.
In a second task version, the robot had to plug a 230\,V CEE-type plug, which was closed with a lid with spring mechanism, into a power outlet at the wall. 
One hand was required to dexterously open the lid and keep it open while the other hand had to insert the plug. 
This task was challenging due to potential self-collisions between hands and due to several occlusions.
For both tasks, the 6D mouse control was well-suited since it allowed very precise motions.
However, force feedback was not present, and contact had to be assessed from visualizations.

In the \emph{Shackle} task, the robot had to fix a shackle to a metal ring at the wall.
One hand had to position the shackle around the ring while the other hand had to insert and turn a screw.
An adapter for the small screw head facilitated grasping and turning.
The 6D mouse interface was used and its precise motions were required to align the screw with the thread.

In the \emph{Electrical Screw Driver} task, the robot had to screw a wooden board to a wooden block by using an off-the-shelf electrical screw driver.
The wooden board with pre-mounted screws was held with the HERI II hand.
The anthropomorphic Schunk hand operated the electrical screw driver using its index finger for triggering.
The 6D mouse control was used to align the tip of the screw driver with the screw head.
When activating the screw driver, the screw moved inside the wood,
requiring fast reaction to follow the screw head towards the wooden board.

Finally, the \emph{Power Drill} task demonstrated the operation of a bi-manual tool by 
making holes at marked places into a wooden block.
Again, the Schunk hand was used to trigger the drill.
The 6D mouse control was set to generate bi-manual motions with respect to a custom frame in the tool.
Since one hole was drilled with a deviation of \texttildelow 1\,cm to the marked position, 
this task achieved a rating of 90\%. 
\begin{figure*}
	\centering
	\scalebox{0.68}{\includegraphics{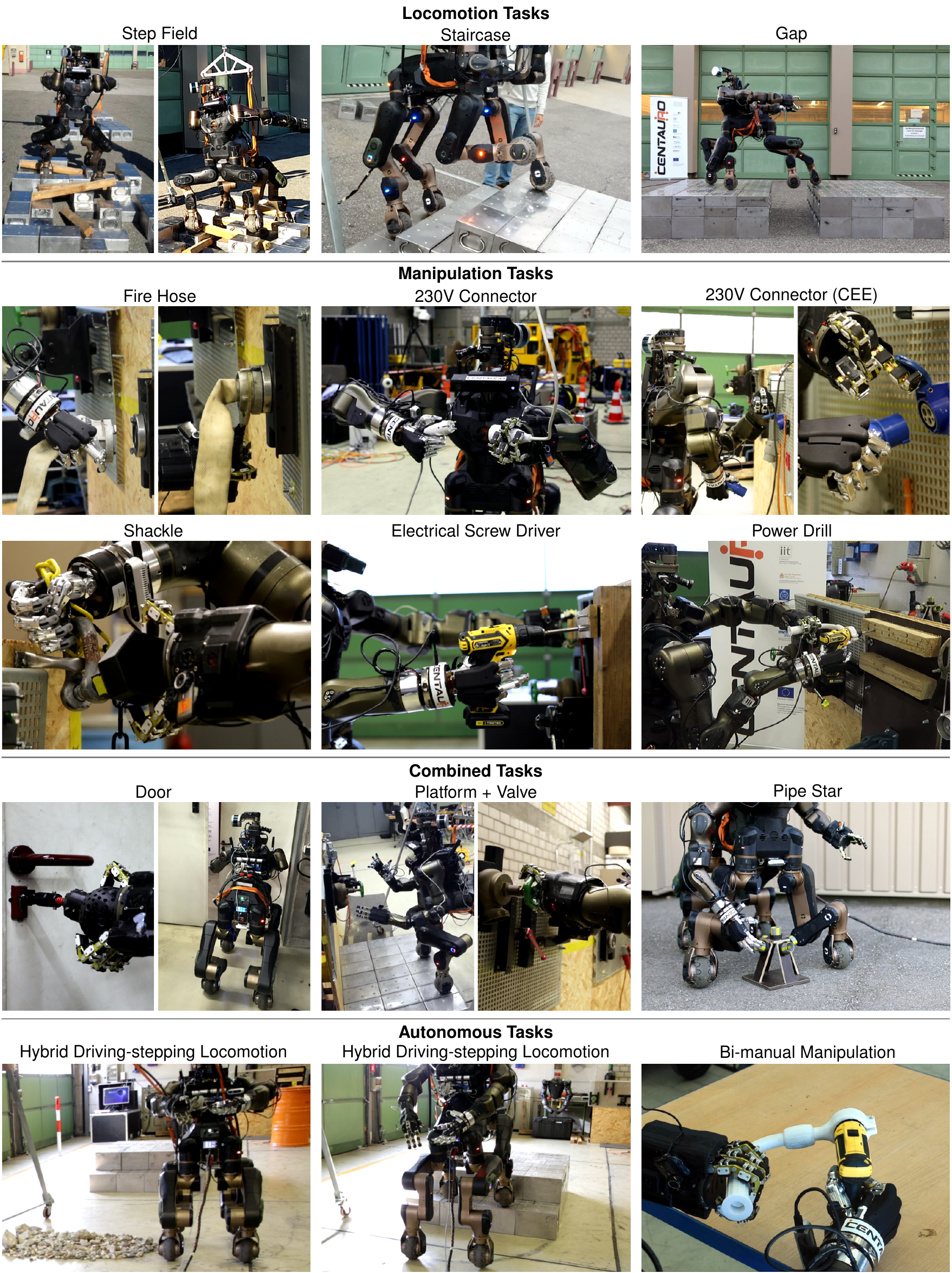}}
	\vspace{-0.2cm}
	\caption{Impressions of the final evaluation at KHG.}
	\vspace{-0.5cm}
	\label{fig:experiments}
\end{figure*}

\subsection{Combined Tasks}

Three tasks combined locomotion and manipulation. 
They were designed to represent realistic missions and to evaluate the coordination of different control interfaces.

In the \emph{Door} task, the robot had to pass a locked door. 
The key was mounted to an adapter which KHG uses in real-world missions.
Once the key was inserted in the lock, multiple re-grasps where necessary to turn it which was realized by 6D mouse control.
After opening the door, the arms were moved to a configuration such that they further opened the door when driving through it.
Omnidirectional driving was controlled by joystick.

In the \emph{Platform\,+\,Valve} task, the robot had to approach an elevated platform, to climb it with its front feet and to open and to close a valve. 
This task was performed with a gate- and lever-type valve.
Approaching the platform was realized by joystick while the semi-autonomous stepping controller was employed to climb the platform.
Centauro's upper body was controlled through the tele-presence suit.
The gate-type valve was turned by 360\textdegree\ in each direction for opening and closing.
While the operators thought that this rotation was sufficient, the referee expected full opening and closing ($>$360\textdegree) such that a rating of only 60\% was given. 
Subsequently, when operating the lever-type valve, a mechanical part broke inside the upper-body exoskeleton after closing the valve for \texttildelow\,30\textdegree. 
Demonstrating the robustness of our system in case of failure, we switched to 6D mouse control \emph{on-the-fly} and finished the task.

Finally, the \emph{Pipe Star} task was performed. 
A pipe star was placed on the ground. 
It consists of five short pipes with different orientations and different symbols inside.
Each of the pipes had to be grasped at the top and the symbol inside the pipe had to be perceived.
This task is used by KHG to evaluate real-world mobile manipulation platforms.
Initially, the keyframe editor was used to move the robot to a very low base pose allowing for manipulation close to the ground (see \cref{fig:experiments}).
Operators used the joystick for omnidirectional driving and the 6D mouse for arm motions.
When grasping the pipes, the top pipe was only touched from the side and not from the top.
While operators thought to have accomplished the task, the referee rated this only as partial success since he interpreted the task objective such that contact from the top was compulsory.
Moreover, it was challenging for the operators to assess physical contact since force feedback was not available.
Visualizing the symbols inside the pipes was realized through the wrist RGB-D camera without problems.

\subsection{Autonomous Tasks}
Autonomous control functionalities were evaluated in two tasks.
In the \emph{Autonomous Locomotion} task, the robot started in front of a patch of debris and some obstacles. 
A two-step staircase ending in an elevated platform was positioned behind the debris with the goal pose on the platform (see \cref{fig:auto_loco_pipeline}).
Map generation, localization, goal pose definition, and planning took around 30\,s.
The robot detected the debris, drove around it, and arrived at the staircase.
It then started to climb the staircase.
However, after climbing two steps with its front feet and by performing the first step with a rear foot, the robot lost balance.
Essentially, the robot model, which was used by the planner, did not include recent hardware modifications (such as integration of the battery) and, hence, based its balance assessment on a wrong center of mass.

In the \emph{Autonomous Manipulation} task, the robot had to detect an unknown, bi-manual power drill on a table in front of it, to grasp it with both hands, to operate the trigger, to lift it, and to put it back on the table.
The robot successfully detected the driller and derived a valid grasping pose which it approached while avoiding collisions with the table.
However, due to a small misalignment of the right hand grasp, it failed to activate the tool and also touched the tool after placing it back on the table such that the tool fell.

\subsection{Isolated Evaluation of Tele-presence Suit Control}

\noindent
The tele-presence suit control was evaluated in isolated experiments at a later meeting.
Again, direct visual contact between the operator and the robot was prevented.
In a first task, a wrench had to be grasped and a valve stem had to be turned.
This task included fine positioning and orientation control of the tool tip, and compliance and modulation of exerted forces on specific axes.
Force feedback enabled the operator to feel the interaction between the wrench and the valve stem facilitating positioning of the tool.

In a second task, bi-manual coordination was evaluated.
The operator had to grasp a glue gun which was positioned in an unfavorable position on a table in front of the robot.
Since a single direct grasp would not allow for correct tool usage, the operator picked up the tool with the HERI II hand, handed it over to the Schunk hand, and adjusted it to obtain a functional grasp which allowed triggering the tool.
This task required precise position and orientation control of both hands combined with adequate grasp forces to avoid slippage.
In addition, problems derived from the force loop where alleviated by the compliance of the controller creating a stable behavior without noticeable oscillations---neither on the operator nor on the robot side.
Task fulfillment took 88\,s.

The tele-presence suit control was fast enough to reliably solve challenging task in feasible time but it was considerably slower than a human executing the task.
Two reasons were identified:
1) the operator received no feedback for collisions between the robot elbow and lower body. 
Introducing another modality such as auditory feedback could help the operator.
2) the robot hands often caused considerable occlusions in the 3D operator visualization when manipulating objects. 
We observed that the main operator did only rely on 3D visualizations if those were available in high quality and with few occlusions. Otherwise, RGB camera images where preferred, but the missing third dimension caused challenges for fast and precise grasping. Fusing RGB-D data from different perspectives could overcome this issue.


\section{Lessons Learned and Conclusion}
\label{sec:conclusion}

\noindent
During the CENTAURO project and its evaluation, systematic integration has played a key role.
Complex sub-systems such as the Centauro robot, the tele-presence suit, or autonomous control functionalities require extensive testing of isolated functionalities.
We have learned to not underestimate the preparation effort a systematic evaluation requires.
Misunderstandings between the operator and the referee occurred since details of individual task objectives were not communicated precisely.

In summary, we presented the final state of the \mbox{CENTAURO} system consisting of the flexible Centauro robot and a set of operator interfaces including a tele-presence suit.
We demonstrated the system capabilities to solve challenging, realistic disaster-response tasks in a wide range.
Our set of operator interfaces was even capable of coping with unexpected events such as broken cameras or the temporal unavailability of individual interfaces.
We further showed that neither pure teleoperation nor pure autonomy are desirable for controlling a complex robot in such challenging tasks. 
Instead, a set of control interfaces which address different tasks and support the operation on different levels of autonomy provides intuitive control and flexibility.

Nevertheless, the system speed to solve tasks is still significantly slower compared to a human. 
We observed that control interfaces with a high degree of autonomy are in general faster in relation to the task complexity. 
Hence, extending the capabilities of autonomous functionalities is promising but requires considerable development effort.
Regarding the speed of direct teleoperation through the tele-presence suit, we discovered that force feedback is valuable as well as other feedback modalities such as vision.
We identified a lack of system understanding as the main reason for the operation speed.

In general, force feedback was found to be valuable to the operator and essential in certain tasks. 
However, it required noticeable hardware complexity of the telepresence suit. 
In future system development, a better trade-off between the performance of the haptic feedback and the complexity of the hardware should be found by exploring different approaches.

Finally, given KHG's extensive experience about challenges of disaster response systems, CENTAURO have overcome several of the described issues including an effective locomotion approach, bimanual manipulation, and operator interfaces which reduce the cognitive load.
In our opinion, the CENTAURO system constitutes a significant step towards the support of rescue workers in real-world disaster-response tasks, eliminating the risks to human health.


\bibliographystyle{IEEEtranN}
\bibliography{paper.bib}

\balance






\end{document}